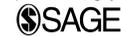

# Identifying High-accuracy Regions in Traffic Camera Images to Enhance the Estimation of Road Traffic Metrics: A Quadtree-based Method

**Yue Lin**[1]  **and Ningchuan Xiao**[1]

## Abstract

The growing number of real-time camera feeds in urban areas has made it possible to provide high-quality traffic data for effective transportation planning, operations, and management. However, deriving reliable traffic metrics from these camera feeds has been a challenge because of the limitations of current vehicle detection techniques, as well as the various camera conditions, such as height and resolution. In this work, a quadtree-based algorithm is developed to continuously partition the image extent until only regions with high detection accuracy remain. These regions are referred to as high-accuracy identification regions (HAIRs) in this paper. We demonstrate how the use of HAIRs can improve the accuracy of traffic density estimates using images from traffic cameras at different heights and resolutions in Central Ohio. Our experiments show that the proposed algorithm can be used to derive robust HAIRs in which vehicle detection accuracy is 41% higher than that in the original image extent. The use of HAIRs also significantly improves the traffic density estimation with an overall decrease of 49% in root mean squared error.

The past two decades have witnessed a surge in world vehicle ownership (*1*) that not only aggravates many negative externalities of the transportation systems, such as road congestion and air pollution in urban areas, but also threatens the sustainability of human–environment systems worldwide (*2*). To address these issues, many countries have started to adopt effective approaches to transportation operations, all of which require accurate and real-time road traffic metrics, such as traffic density, flow, and speed. In transport and urban planning, road traffic metrics offer critical data support to the arrangement of transport services and infrastructure that facilitates equitable access to employment, system reliability, and social/environmental justice (*3*). Today, road traffic metrics are widely applied in various areas, including routing (*4*), dynamic congestion pricing (*5, 6*), traffic emission control (*7, 8*), and intelligent transportation systems (*9*).

Since the early 2000s, traffic cameras have rapidly emerged as a primary data source in the field of transportation, particularly in the U.S.A. and Canada. Real-time images from these cameras are typically available to the public. The April 2020 release of the Traffic Video Mobile App by TrafficLand (*10*), for example, provides access to more than 25,000 traffic cameras in over 200 cities in the U.S.A. for the public to use through free subscriptions. These publicly available camera feeds have the potential of being used to provide accurate and real-time road traffic metrics at scale with relatively low additional capital costs. Many vehicle detection methods have been developed to process these camera images for road traffic metric estimation, and significant progress has been made in this area since several deep learning approaches were proposed (see a review in the next section). However, vehicles in the same images are not equally detectable. As indicated by Zhang et al. (*11*), objects in the foreground of an image can generally be detected with high accuracy by most object detection methods, whereas those distant from the camera because of tilt and perspective are often too small to be accurately detected. Missed and/or falsely identified small vehicles tend to result in significant errors in the estimation of road traffic metrics.

[1]Department of Geography, The Ohio State University, Columbus, OH

**Corresponding Author:**
Yue Lin, lin.3326@osu.edu

The purpose of this paper is to develop an effective algorithm that can be used to achieve accurate traffic metric estimation from camera feeds. We address the shortcomings of existing methods with a quadtree-based algorithm that recursively partitions the image so that only the high-accuracy regions are used in traffic metric estimation. These regions are referred to as the high-accuracy identification regions (HAIRs), and focusing on them can lead to a more reliable and accurate estimation of road traffic metrics, such as traffic density. In the remainder of this paper, the second section discusses the rationale of this research by examining the limitations of existing data sources and vehicle detection methods. The third section presents our algorithm for HAIR identification that helps improve the accuracy of the road traffic metrics derived from camera images. In the fourth section, the proposed algorithm is applied to the traffic camera data collected in Central Ohio for accurate traffic density estimation. The paper is concluded in the fifth section.

## Background and Rationale

Road traffic metrics can be obtained from different data sources. For example, vehicular networks formed by vehicle-to-vehicle (V2V) communication technology have been used to provide road traffic metrics (*12, 13*). This approach relies on participating private or commercial vehicles with built-in global positioning system (GPS) trackers, although recruiting participants can be time-consuming and costly. Other methods use data from mobile phone tracking, which are often restricted by device usages and may be subject to privacy concerns about the identities and private movement behaviors of users (*14*). Data sets developed based on V2V or mobile phone tracking are generally not accessible to the public, which also limits their applications in transportation. Another type of method utilizes traffic sensors, including loop detectors, pneumatic tubes, and light detection and ranging (LiDAR) sensors (*15, 16*). However, these sensors often suffer from relatively short lifetimes, high capital costs in maintenance, and limited spatial coverage, and therefore have not been prevalently adopted for robust traffic monitoring (*17*).

Traffic camera feeds are an emerging data source that is publicly accessible and has a wide coverage. Vehicle detection is an essential step to extract road traffic information from camera feeds. Traditional vehicle detection methods utilize classical computer vision techniques, such as edge and corner detection, to identify the presence of individual vehicles from images (*18, 19*). However, these methods are based on low-level vehicle features (e.g., straight lines and dots on the vehicle) that are small and simple, which often leads to a considerable number of missing and falsely detected vehicles in the images (*20*). Over the past decade, the rapid development of deep learning methods, especially the recent breakthrough of convolutional neural networks (CNNs), has significantly improved the accuracy of many computer vision applications, including vehicle identification (*18, 21*). The CNN is a class of deep neural networks that can automatically detect and "learn" high-level features representing larger and more complex vehicle elements (e.g., wheels of a vehicle), which, compared to traditional computer vision methods, can utilize more informative and representative features to detect vehicles in the images (*20*). To date, a set of CNN-based object detectors has been developed and applied to vehicle detection, including Faster R-CNN (*22*), YOLO v4 (*23*), RetinaNet (*24*), and EfficientDet (*25*).

Existing deep learning detectors heavily depend on the level of detail of the vehicles present in the images to make accurate detections. As indicated by Wang et al. (*26*), for vehicles with ground truth bounding box sizes larger than 40 pixels, the average precision in detection can generally reach above 0.8, because details of the vehicle element and feature are clearly displayed in the images. When the vehicle sizes are below 25 pixels, the average precision drops to approximately 0.4, implying that many vehicles are missed and/or falsely identified. Figure 1 presents two examples of using a deep learning detector to identify the vehicles in traffic images (see the Appendix for details about the detector). Because of the tilts and perspectives of traffic cameras, the level of detail of the on-road vehicles can vary significantly in the same image. Vehicles distant from the cameras can be of small size and even not identifiable with the naked eye. These distant and small vehicles in the image lack detailed vehicle features and therefore are unlikely to be correctly identified by vehicle detectors (*27, 28*).

The tendency of vehicle detectors in misclassifying distant vehicles in images leads to a dilemma when we use the detected vehicles to compute road traffic metrics: should we use the entire image or just the part in which detected vehicles are presented in the image? For example, calculating traffic density in an area relies on accurate measures of both vehicle count and road length. It is inappropriate to use the entire image as the area because the vehicles in the distant areas from the camera are almost certain not to be "counted" by the vehicle detector used (note that all vehicle detection methods have similar performance over these vehicles), and subsequently dividing the vehicle count by the road length in the entire image extent is erroneous. Fortunately, this threat to internal validity (*29*) can be effectively addressed by simply excluding the areas in which errors tend to occur. In other words, we establish a region formed by a subset of the pixels in the image in which we can consistently detect vehicles accurately. We call these regions the HAIRs. Vehicles appearing outside the HAIRs, detected or not, will not be counted. In the same fashion, only roads within the HAIR will be used for traffic density (or other metrics, such as speed) calculation.

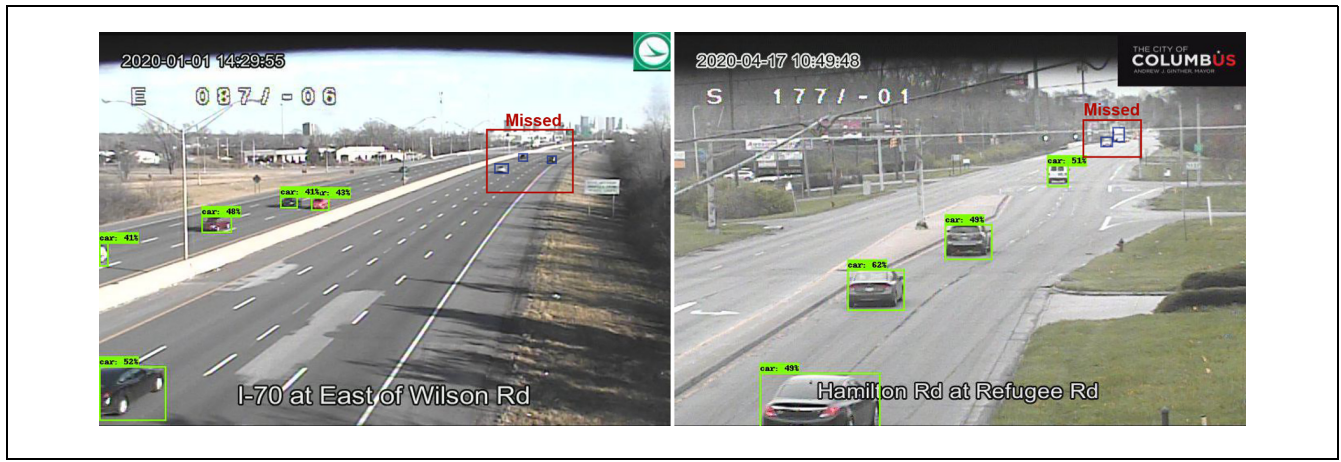

**Figure 1.** Two examples of using EfficientDet (25) to identify vehicles in traffic images. Rectangles with blue outlines are the ground truth bounding boxes of missed vehicles, and those with green outlines are the bounding boxes of vehicles predicted by the model. (Color online only.)

## Quadtree-Based Method for Hair Identification

Three important factors need to be considered when constructing the HAIR: accuracy, consistency, and efficiency. Firstly, a high and desired vehicle detection accuracy, as specified by the user, should be reached within the HAIR. Although there may not exist an objective accuracy level, users on different applications often have their desirable accuracy, and the algorithm should be able to identify regions that exceed such an accuracy threshold. Secondly, we should expect the HAIR to perform consistently for new (existing or future) images with the same field of view. Finally, the calculation of the HAIR should be efficient as it should not add computational burdens to the vehicle detection process.

These three requirements help us explore the use of image segmentation algorithms in the field of computer vision for HAIR identification. Image segmentation refers to the process of continuously partitioning an image into regions until the regions become homogeneous (30). Among the many methods developed for image segmentation, such as pixel-wise masking methods of region growing or data clustering (31, 32), the region quadtree-based method (33, 34) has been extensively applied and proven to be efficient and capable of producing high-quality results (35–38). In the region quadtree-based algorithm, an image is divided into four equal quadrants, and if a quadrant is not homogeneous, it will be further decomposed into smaller quadrants. Figure 2 illustrates such a process. A parameter called depth is used to describe the level of partitioning, and the depths for the three trees in Figure 2, *a–c*, are 0, 1, and 2, respectively.

When the quadtree-based partitioning process is applied to different images (of the same size), the actual partition may vary, but the process and data structure will remain the same. For example, all partitions will start by splitting the images into four quadrants of the same size. This feature is particularly suitable for this study because it provides the necessary consistency across multiple images. We adopt the quadtree-based segmentation approach to partition traffic camera images so that consistent high-accuracy regions can be established for traffic metric calculation. Figure 3 presents a flowchart for implementing our quadtree-based algorithm. A vehicle detector is first used to produce a collection of traffic images with detected vehicles. Based on the detection and a set of user-defined parameters, we run the quadtree-based algorithm that returns the HAIR for a camera. The HAIR and the detected vehicles will then be used for traffic metric estimation.

### Parameters

The purpose of our algorithm is to determine the regions in a camera's field of view to be the HAIR. This algorithm requires three user-specified inputs: number of images, accuracy threshold, and maximum depth. Firstly, a set of images is needed in which each image contains (1) the manually labeled vehicle bounding boxes that are confirmed to be the ground truth of the vehicles to be detected, and (2) the bounding box of each detected vehicle yielded by some model. We use $N$ to denote the number of images.

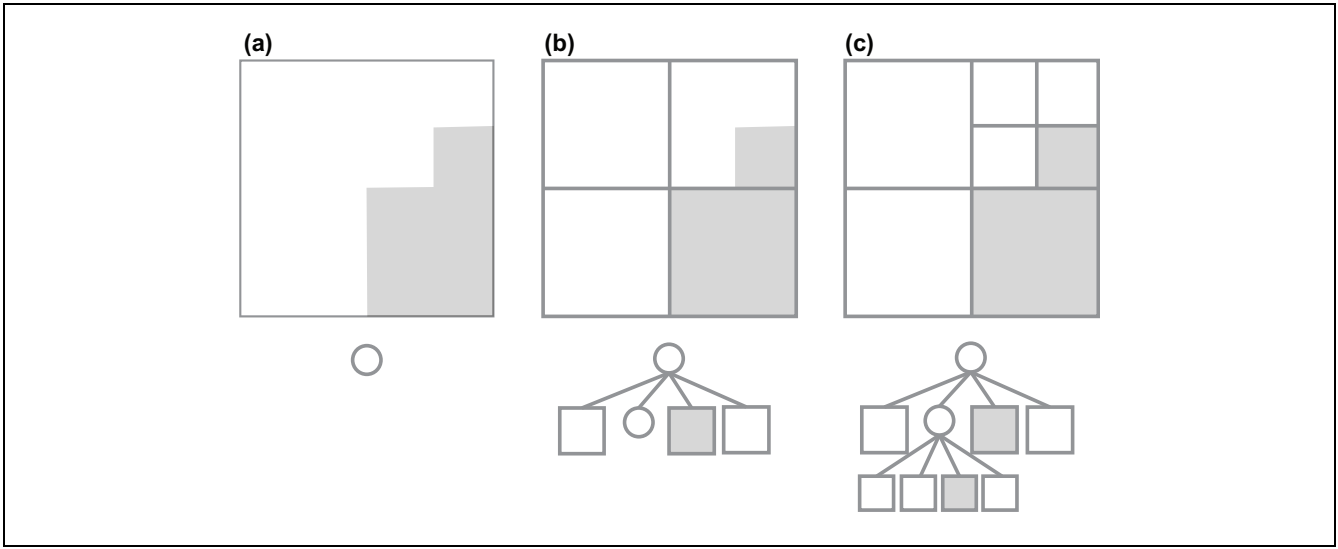

**Figure 2.** A region quadtree-based image segmentation process. The image (upper row) to be segmented and the corresponding quadtree representations (lower row) are presented. In the tree, a circle represents a heterogeneous quadrant that can be further partitioned before reaching the maximum depth, and a square represents a homogeneous quadrant. The heterogeneous image: (*a*) is divided into four equal quadrants (*b*), in which northwest, northeast, southeast, and southwest quadrants are represented in the four branches (from left to right) of the tree underneath (*b*), respectively. The quadtree for (*b*) has a depth of 1. The upper-right quadrant in (*b*) is heterogeneous, as represented by the circle in the second branch of the tree, and needs to be further partitioned into smaller quadrants (*c*). The partitioning stops if all quadrants are homogeneous or a maximum depth is reached.

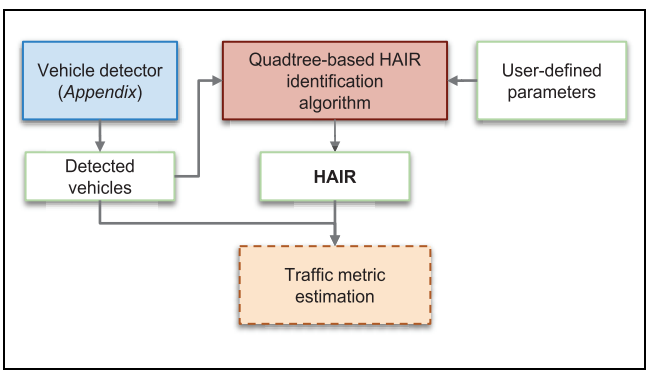

**Figure 3.** A flowchart of the quadtree-based method.
*Note:* HAIR = high-accuracy identification region.

The second parameter relates to the accuracy we aim to achieve by identifying the regions of the image as the HAIR. In our process, the image is partitioned into four equal quadrants and the partitioning of a quadrant stops when the accuracy *within* the quadrant is higher than a user-specified threshold $a_0$. To evaluate the vehicle detection accuracy of a given region (a quadrant, the entire image, or any arbitrary region), we first compile the bounding boxes of all the vehicles in the region. We ensure that each vehicle (detected or labeled) belongs to one and only one region. If a vehicle's bounding box intersects with multiple regions, the vehicle will be assigned to the region with the largest overlapping area.

The bounding box of each detected vehicle $B_d$ is compared to that of its closest labeled vehicle $B_{gt}$ to determine whether the detection is correct. A metric called intersection over union (IoU) is calculated as the ratio between the intersected area of $B_d$ and $B_{gt}$ and their union area:

$$\text{IoU} = \frac{\text{area}\left(B_d \cap B_{gt}\right)}{\text{area}\left(B_d \cup B_{gt}\right)} \qquad (1)$$

When the IoU value exceeds 0.5 (or 50%), more than half of the detected bounding box overlaps with a vehicle that is considered as its ground truth, and it is reasonable to say that this vehicle is correctly detected (*39*). A correctly detected vehicle is a true positive (TP). A detected vehicle with an IoU value smaller than 0.5 is a false positive (FP) because we cannot find a vehicle in the ground truth to match it. Ground truth vehicles that are not detected by the model are false negatives (FNs).

For a given region, we define a metric called the regional average precision (RAP) to evaluate its vehicle detection accuracy. RAP is similar to the average precision in the literature (*39*). Each detected vehicle is associated with a confidence score reported by the vehicle detector to indicate its predicted probability of being a vehicle object. Based on the confidence scores, the list of the detected vehicles in the region is sorted in descending order. For the top *i* objects in the list, we have a pair of values to evaluate the detection accuracy: precision and recall. Recall is the proportion of TPs among *all* ground truth vehicles (i.e., TPs + FNs) within the region, calculated as follows:

$$\text{Recall} = \frac{\text{TP}}{\text{TP} + \text{FN}} \qquad (2)$$

Precision is the proportion of TPs among the detected vehicles (i.e., TPs + FPs) within the top $i$ objects in the list, calculated as follows:

$$\text{Precision} = \frac{\text{TP}}{\text{TP} + \text{FP}} \qquad (3)$$

Let $R = \{r_1, r_2, \ldots, r_M\}$ be a sequence of $M$ recall levels with $r_1 = 0$, $r_M = 1$, and $r_k \leqslant r_{k+1}$ ($1 \leqslant k < M$), and $p(r_k) = \max\{p(r_j), j \geqslant k\}$ the maximum precision value at recall level $r_k$ (i.e., the maximum precision associated with recall values exceeding $r_k$). The RAP metric is the average of the maximum precision values at all recall levels in $R$:

$$\text{RAP} = \frac{1}{M} \sum_{r_k \in R} p(r_k) \qquad (4)$$

A common practice to define the recall levels $R$ in the literature (*39*) is to use $M = 11$ equally spaced values such that $R = \{0, 0.1, \ldots, 1\}$. This is also adopted in this paper.

Figure 4 presents a hypothetical example of how the RAP of a region can be calculated when we have multiple images from a camera. In this case, the entire image is the region. Within each image, the detected vehicles are classified as either TPs or FPs, and all missed vehicles (FNs) are also marked. In this example, the RAP value is 0.55. In a practical sense, a high RAP score indicates a high probability of vehicles in a given region being correctly detected. It can be observed that if the region is partitioned into four quadrants, both the lower-left and lower-right quadrants may have higher RAP values (more on this in the next section).

The third parameter is the maximum depth $d_0$ used in the algorithm to control the level of partitioning. Even though the partitioning continues until the vehicle detection accuracy of a quadrant reaches the accuracy threshold $d_0$, theoretically each image can be partitioned to the level of a single pixel, which is unnecessary because vehicle bounding boxes are always larger than that.

## Algorithm

The quadtree-based HAIR identification algorithm is a recursive algorithm that consists of the steps outlined below, in which $I$ denotes the extent of the region (entire image or a quadrant at any depth) to be examined, $H$ a list to store the quadrants that form the HAIR, and $d$ is the current depth.

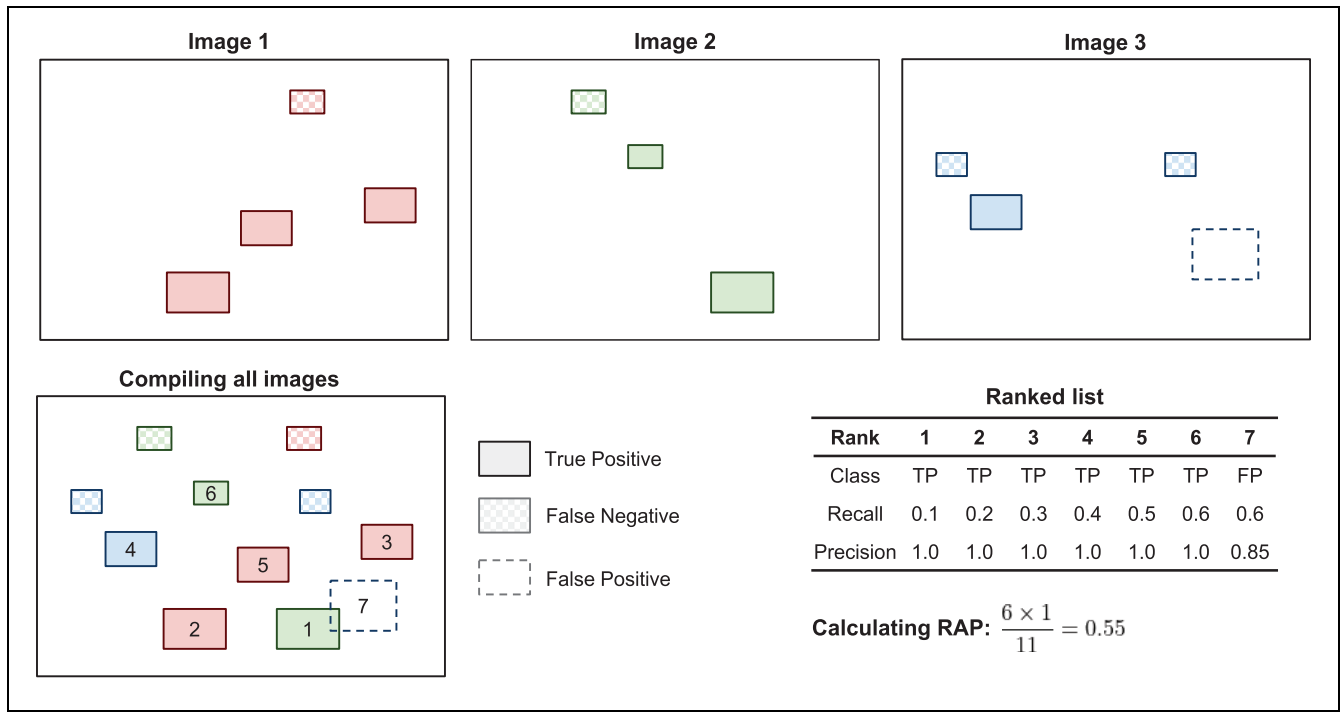

**Figure 4.** An example of calculating the regional average precision (RAP) of a hypothetical region using three images from a camera. Numbers in the compiled region are the hypothetical ranks of confidence score for each detected vehicle, reflecting the trend that vehicles close to the camera tend to have high confidence scores. The maximum precision values are 1 at six recall levels from 0.1 to 0.6, respectively, whereas the rest are 0, and therefore the RAP value is 0.55.
*Note:* TP = true positive; FP = false positive.

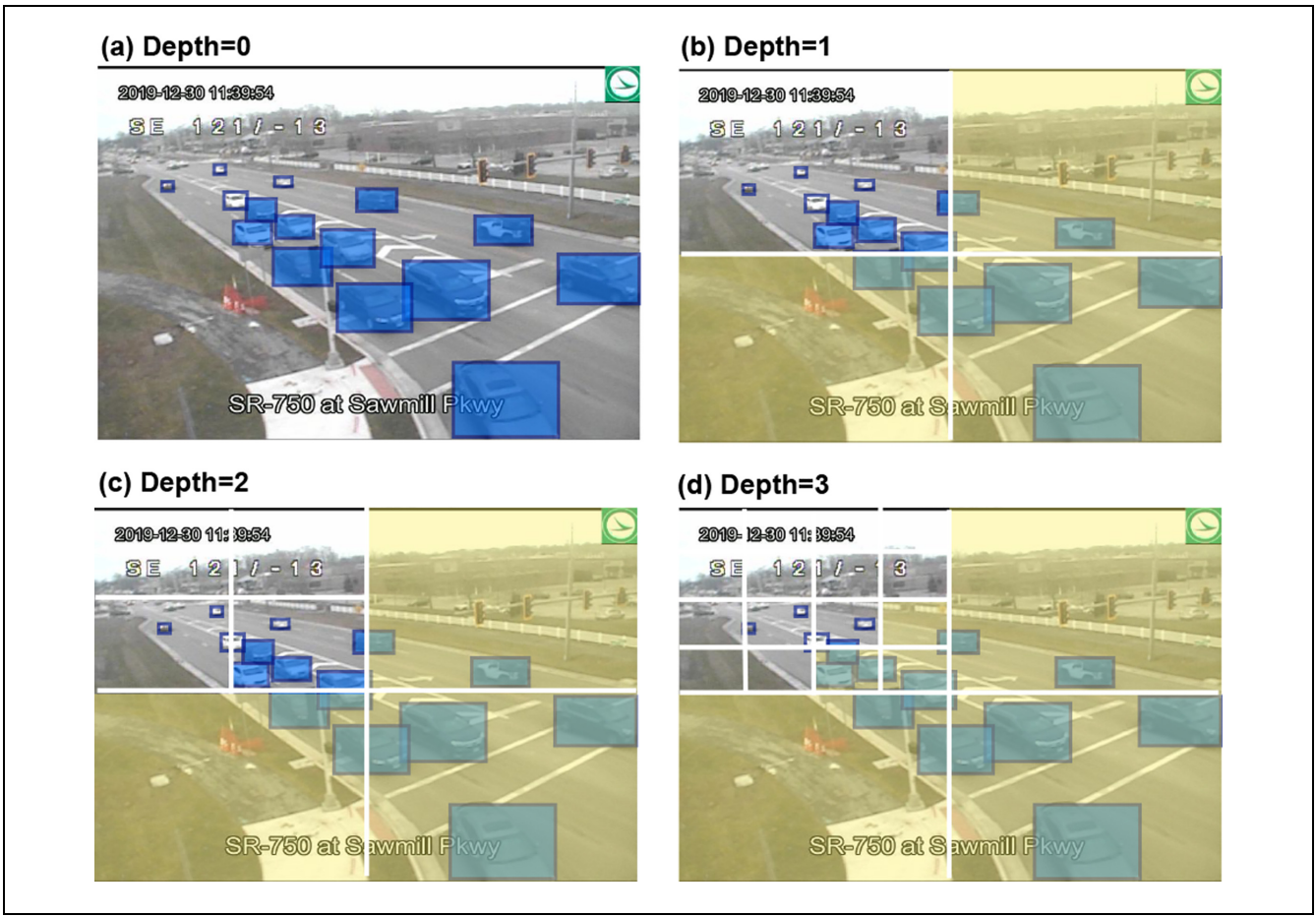

**Figure 5.** An example of using the quadtree-based algorithm for high-accuracy identification region (HAIR) identification. Regions rendered in yellow are parts of the HAIR obtained at different depths. Rectangles with blue outlines are the bounding boxes of annotated vehicles, in which the filled blue rectangles are vehicles that are correctly detected (true positives) and the open ones are those not detected (false negatives). (Color online only.)

**Step 1.** Set $I$ to the entire image extent, $H$ to an empty list, and $d$ to 0.
**Step 2.** Combine all the detected and labeled vehicles that fall within $I$ in the $N$ images, and compute the RAP of $I$ using Equation 4. If the RAP is greater than $a_0$, add $I$ to $H$ and stop.
**Step 3.** If $d$ equals $d_0$, stop.
**Step 4.** Partition the image into four equal quadrants and increase $d$ by 1.
**Step 5.** For each quadrant, set $I$ to the extent of the quadrant and repeat from Step 2.

Figure 5 illustrates the quadtree-based HAIR identification algorithm using one traffic image. For demonstration, we set the accuracy threshold $a_0$ to 0.75, and the maximum depth $d_0$ to 3. Without showing the detailed calculation, we note that the RAP value of the original image extent (Figure 5a) is 0.727. This is lower than $a_0$ and therefore the image will be partitioned into four equal quadrants (Figure 5b). At depth 1, the southwest (lower-left), southeast (lower-right), and northeast (upper-right) quadrants now have a RAP value of 1, and they will be included in the HAIR (rendered in red) and will not be further partitioned. The other quadrant, however, has a RAP of 0.45, which is lower than 0.75, and the partitioning continues. Figure 5c shows the result of the partitioning at depth 2, in which all four new quadrants are still below the threshold. After the next round of partitioning (depth 3), three additional quadrants at this depth exhibit a RAP higher than the threshold $a_0$ and are then added into the HAIR, whereas others still have low RAP values (Figure 5d). As the current depth is equal to $d_0$, the partitioning process stops at this level and the HAIR for this image is finalized. In some cases, further decomposing the image may lead to more regions

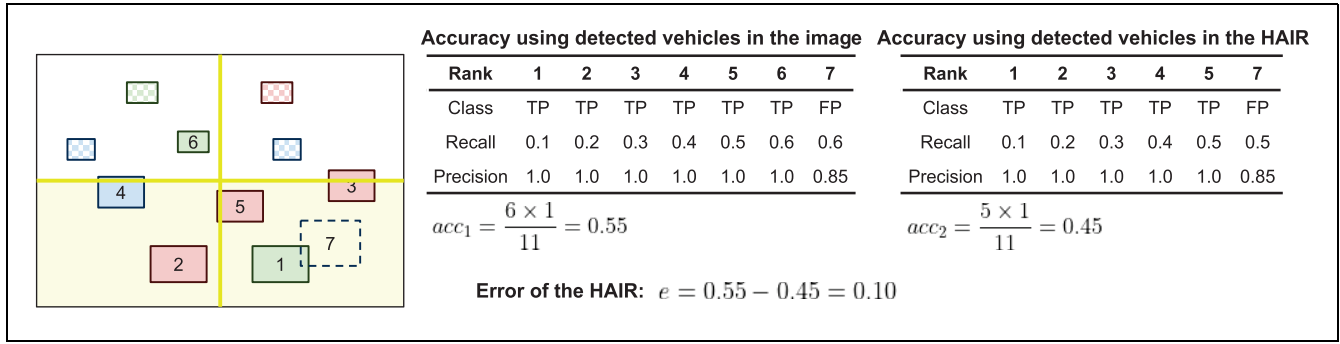

The table in the figure:

**Accuracy using detected vehicles in the image**

| Rank | 1 | 2 | 3 | 4 | 5 | 6 | 7 |
|---|---|---|---|---|---|---|---|
| Class | TP | TP | TP | TP | TP | TP | FP |
| Recall | 0.1 | 0.2 | 0.3 | 0.4 | 0.5 | 0.6 | 0.6 |
| Precision | 1.0 | 1.0 | 1.0 | 1.0 | 1.0 | 1.0 | 0.85 |

$$acc_1 = \frac{6 \times 1}{11} = 0.55$$

**Accuracy using detected vehicles in the HAIR**

| Rank | 1 | 2 | 3 | 4 | 5 | 6 |
|---|---|---|---|---|---|---|
| Class | TP | TP | TP | TP | TP | FP |
| Recall | 0.1 | 0.2 | 0.3 | 0.4 | 0.5 | 0.5 |
| Precision | 1.0 | 1.0 | 1.0 | 1.0 | 1.0 | 0.85 |

$$acc_2 = \frac{5 \times 1}{11} = 0.45$$

**Error of the HAIR:** $e = 0.55 - 0.45 = 0.10$

**Figure 6.** An example of calculating the error brought about by the use of the high-accuracy identification region (HAIR). The region here denotes the image extent, and the HAIR includes the two lower quadrants (in yellow). All the correct detections fall in the HAIR except the one ranked sixth. (Color online only.)
*Note:* TP = true positive; FP = false positive.

being included in the HAIR. However, it should be noted that such decomposition may become meaningless if the resulting sub-quadrants are too small to contain vehicle bounding boxes. We will examine the impact of depth in the fourth section in which computational experiments are discussed.

At this point, one may wonder whether the use of the HAIR can improve the accuracy of vehicle detection. It is important to clarify that the HAIR does not improve vehicle detection accuracy per se. The HAIR is designed to improve the accuracy with which road traffic metrics such as traffic density are estimated. The example in Figure 5 demonstrates this point. Because many of the vehicles in the upper part of the image are not counted, calculating the traffic density by dividing the number of detected vehicles (7) by the *total* road length within the image tends to underestimate the density. The estimation can be improved by focusing only on the part of the image in which the vehicles are more correctly detected. In this example, dividing the number of vehicles by the road length *in the lower part of the image* (i.e., the HAIR) yields a more accurate estimate of traffic density. When estimating the road traffic metrics, the use of the HAIR allows us to focus only on the areas in which vehicles are correctly detected and exclude those areas in which we cannot correctly detect them. We will further elaborate on this issue in the next section.

### Error Measurement

The HAIR is designed to focus on the regions in an image in which vehicles are accurately detected. The ultimate error of using the HAIR arises when vehicles that can be accurately detected are not included in the HAIR. Such an error can be introduced in two situations. The first situation occurs when the images used are not representative in covering sufficient on-road vehicle positions. A HAIR derived using an insufficient number of images, for example, may cause the error because vehicles on new images may appear at locations that have not been considered in the algorithm. The error may also occur in a situation when a small value of depth is used. For example, in Figure 5, many correctly detected vehicles will be excluded from the HAIR if the process stops at depth 2.

We define the error of the HAIR, $e$, as the loss of accuracy caused by only including the vehicles in the HAIR. In Figure 6, $acc_1$ is the RAP value calculated using all the detected vehicles, which is 0.55. We can then calculate another RAP value, $acc_2$, using only the detected vehicles that fall inside the HAIR, and in this case, all but the one marked as rank 6 are included. If there are correctly detected vehicles outside the HAIR, $acc_2$ will be lower than $acc_1$. In our example here, the error is 0.1. In an ideal situation, all detected vehicles are in the HAIR, and $acc_2$ will be as same as $acc_1$, meaning the error is zero and the use of the HAIR does not induce error.

## Computational Experiments and Results

To demonstrate the effectiveness of the HAIR identification algorithm, we apply it to the images from six traffic cameras in the metropolitan area around Columbus, Ohio, to improve the accuracy of traffic density estimation. These cameras are deployed by the Ohio Department of Transportation (ODOT), with different mounting heights and resolutions. Their configuration details, along with an example image from each of them, are presented in Figure 7.

The use of the HAIR assumes that some object detector is available to identify the vehicles present in the images, and an image data set can be developed in which vehicles are labeled and detected. The algorithm for HAIR identification does not depend on the specific vehicle detector. We train and test a deep learning model called EfficientDet (25) to detect the vehicles in the images. Details of the training and testing are discussed in the Appendix. Then, we collect a random sample of 110 images for each camera during the daytime in December 2019, in which all the motor vehicles present in the images are manually identified and labeled as the ground truths. The number of images used here (110) is determined based on multiple preliminary trials, and we further discuss the impact of image numbers in the *HAIR Identification and Assessment* section. The best detector trained and tested is used to detect the vehicles in the 110 images of each camera. Based on the labeled ground truth and detected vehicles, we conduct two sets of experiments. The first is to identify the HAIR for each camera, and the second is to estimate and assess traffic density in the HAIR.

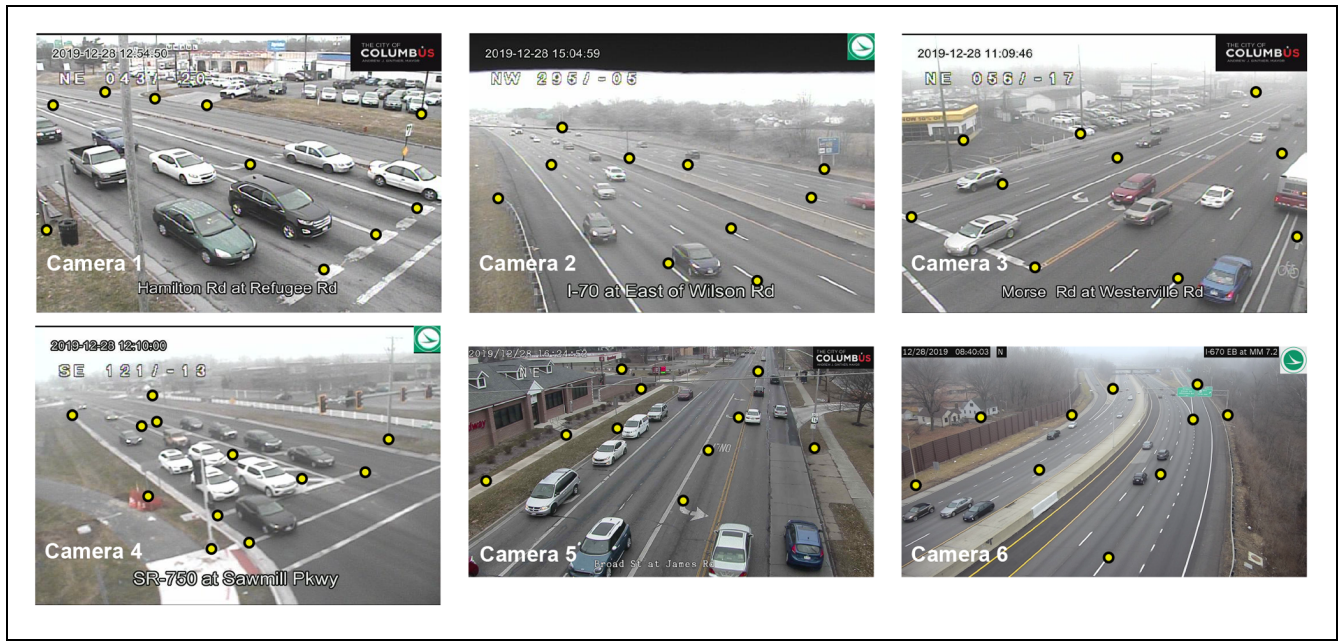

**Figure 7.** Sample images taken by the six selected traffic cameras. Cameras 1, 3, 4, and 5 are mounted low along local roads, each capturing a road section of approximately 420 ft within its field of view. Cameras 2 and 6 are mounted high on highways, and each captures over 0.25 mi of the road section. In addition, cameras 1–4 are low resolution at 704 × 480 pixels, whereas cameras 5 and 6 have a higher resolution of 1920 × 1080 pixels. Yellow dots on each image are ground control points used for georeferencing, after which the length of road sections on the image can be calculated. (Color online only.)

### HAIR Identification and Assessment

A resampling procedure (40) is adopted to find the parameter values for the quadtree-based algorithm so that the average error brought about by the HAIR is minimized for each camera. Here, $N$ images are randomly sampled from the 110 images of each camera. These $N$ images are used to obtain the HAIR for each camera. The value of $N$ is systematically changed from 10 to 100 with an interval of 10, and at each $N$, the maximum depth $d_0$ is systematically changed from 1 to 5. We adopt a RAP threshold $a_0$ of 0.75 that is considered to be a high accuracy in vehicle detection within the given region (39). The error of the HAIR ($e$, see the *Algorithm* section) obtained for each combination of $N$ and $d_0$ is evaluated using 10 images sampled from the remaining 110-$N$ images. We repeat this process (sampling $N$ images, obtaining the HAIR, computing the error using 10 random images) 1000 times and compute an overall error as the root mean square error, or RMSE in short (41):

$$\text{RMSE} = \sqrt{\frac{\sum_{i=1}^{1000} e_i^2}{1000}} \qquad (5)$$

where $e_i$ is the error brought about by the use of the HAIR in the $i$th iteration. By systematically changing the values of $N$ as well as the maximum depth, $d_0$, the trend of the RMSE values is then used to identify a robust final HAIR that can be applied to derive accurate road traffic metrics using new camera feeds.

We expect the RMSE to decrease and then converge when $N$ and $d_0$ increase, a trend that can be used to determine the parameters to derive the final HAIR. The RMSE values for the six cameras under different parameter settings are shown in Figure 8. It can be observed that the RMSE values tend to decrease and then stay at a low value when $N$ and $d_0$ increase. Specifically, increasing the maximum depth $d_0$ while fixing the number of

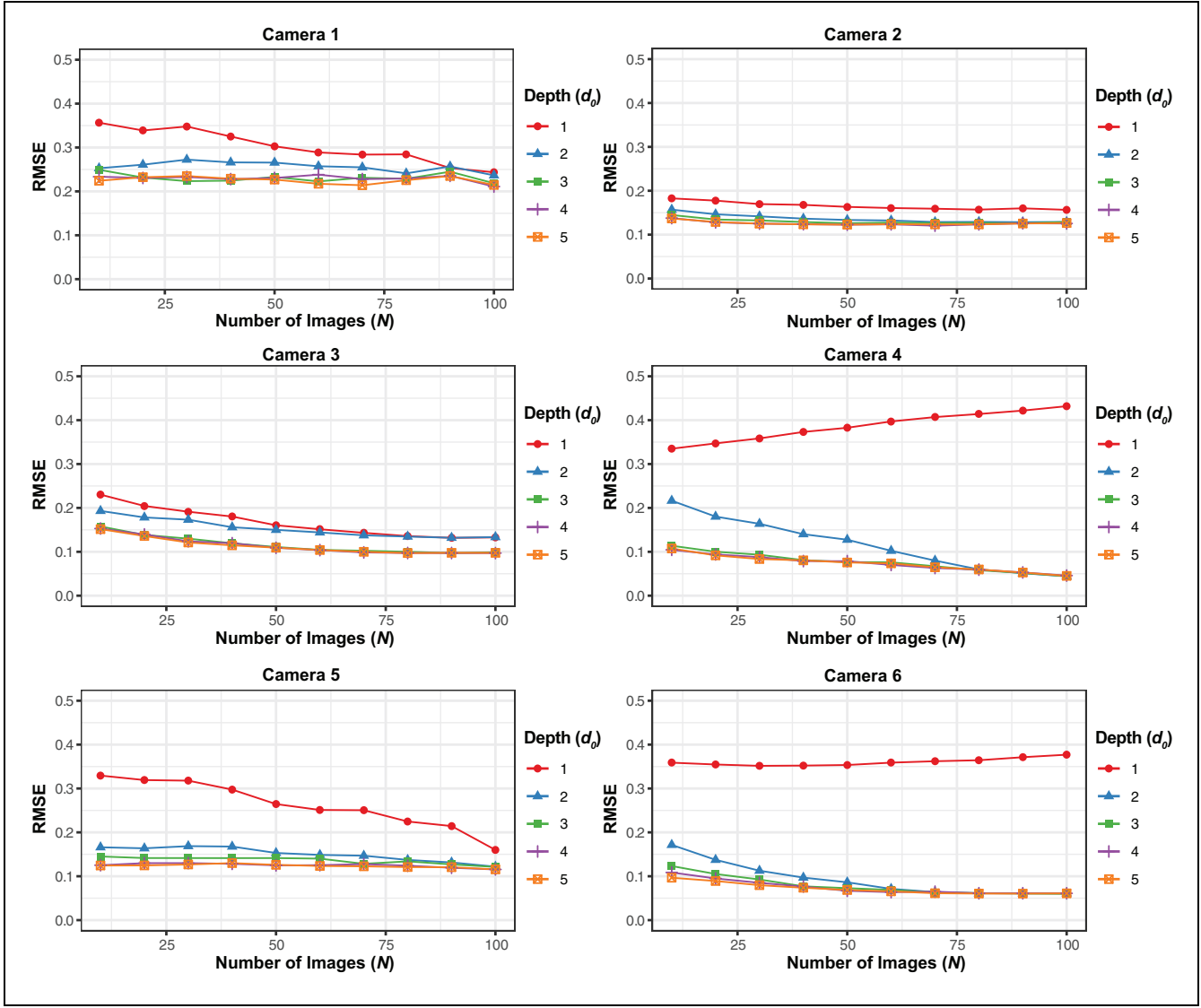

**Figure 8.** The average error of the high-accuracy identification region (HAIR) at different numbers of images $N$ and maximum depths $d_0$ for the selected six cameras.
RMSE = root mean square error.

images $N$ leads to a decrease in the RMSE. This trend suggests that further dividing the images can help discover additional (and small) areas where vehicles can be correctly detected and thus contribute to improving the coverage of the HAIR to include more correctly detected vehicles within the entire image extent. However, when the $d_0$ is too high (larger than 4 in our case), the RMSE remains approximately unchanged in further subdivisions. This is because further subdividing the images will produce the sub-quadrants that are too small to cover new correctly detected vehicles in the images, which are unlikely to be considered for the HAIR. On the other hand, when the maximum depth $d_0$ is fixed, increasing the number of images $N$ to derive the HAIR contributes to the decrease in RMSE. The decrease diminishes when

$N$ reaches a certain level, showing a trend of convergence in the error of the HAIR and suggesting it is not necessary to continuously increase $N$ to reach a robust coverage of detected vehicles in the HAIR. There are two exceptions when $d_0$ is 1 for cameras 4 and 6, but based on the previous discussion about the effect of $d_0$, all our experiments show that $d_0$ needs to be higher than 1 to achieve the accuracy threshold.

The $N$ and $d_0$ used to identify the final HAIR for each camera can be determined based on the RMSE trends (Figure 8). The decrease in RMSE is defined as $\Delta_{ij}$ as the depth increases from $i$ to $i + 1$ with the number of images being $j$ ($1 \leqslant i < 5$ and $1 \leqslant j \leqslant N$, in which 5 is the maximum depth as used in this study). The final maximum depth $d_0^*$ is the smallest depth at which additional increases in depth will not result in substantial accuracy gains. This is determined as the minimum $i$ that satisfies $\Delta_{ij} \leqslant 0.01$ for all $j$, in which the value of 0.01 is a threshold representing a decrease in RMSE without significant gains in accuracy. Once $d_0^*$ is derived, we determine the final number of images $N^*$ for a given camera as the minimum $j$ that satisfies $\Delta\ d_0 * j \leqslant 0.001$. Table 1 lists the final $N^*$ and $d_0^*$ for the six cameras. We also present the average runtime for identifying the HAIR in Google Colaboratory, a free online Jupyter notebook environment that provides one Xeon CPU @ 2.20 GHz for our experiments.

**Table 1.** Final Values for the Number of Images $N^*$ and Maximum Depth $d_0^*$ for the Six Cameras, as Well as the Average Runtime of Identifying the HAIR Using the Algorithm in the *Algorithm* section

| Camera | $N^*$ | $d_0^*$ | Runtime (s) |
|---|---|---|---|
| 1 | 40 | 4 | 22.59 |
| 2 | 20 | 3 | 4.64 |
| 3 | 50 | 3 | 9.01 |
| 4 | 50 | 3 | 4.30 |
| 5 | 30 | 4 | 0.56 |
| 6 | 40 | 3 | 4.27 |

We use 50 annotated images (in which vehicles are identified and labeled) collected from the same six cameras to evaluate the effectiveness of the HAIR, in which the images are not used in the identification of the HAIR. Table 2 presents the detection accuracies for both the original image extent and the HAIR, as well as the average runtime to count vehicles within the HAIR using the quadtree structure in Google Colaboratory. The precision here is the ratio of correctly detected vehicles to all detected vehicles, and the recall is the ratio of correctly detected vehicles to all labeled vehicles, within the extent of the image or the HAIR. The RAP for the image and the HAIR are calculated as described in the *Parameters* section. The overall precision of the entire image extent is generally high for all six cameras (0.972), but the overall recall values are relatively low (0.588 on average), especially for cameras 2 and 3. This means that although more than 90% of the detections by the trained model are correct, a significant number of vehicles are still not identified, which also leads to low RAP values (0.577 on average). The results clearly indicate that the overall recall and RAP inside the HAIR are significantly higher than those of the original image extent, with 48% and 41% increases, respectively. For cameras 2 and 3, the RAP within the HAIR reaches above 0.7, increases of 168% and 98%, respectively, over that within the image extent.

## Traffic Density Estimation

To assess the effectiveness of the HAIR in traffic density estimation, we measure the predicted and observed traffic density values. For both extents of the entire image and the HAIR, the observed traffic density is calculated using several labeled vehicles and the road length with the extent, and the predicted traffic density is obtained using the detected vehicles only. Again, we use the 50 images mentioned above for each camera. For each image, the difference between the observed and predicted values is estimated, and then the root mean square of the differences for the 50 images is calculated as the overall error. The errors for the entire image and the HAIR are presented in Table 3. It is clear that our proposed method using the HAIR can provide more accurate density estimates than those based on the original image extent, reducing overall error by 49%. For camera 3, the error is extremely high (9.939) if we directly use the vehicle count from the entire image. Our method significantly reduces the error to 3.539, a 64% improvement.

## Discussion and Conclusions

This paper focuses on the areas within a camera view where vehicles can be correctly detected to enhance the road traffic estimates, and a quadtree-based algorithm to

**Table 2.** Comparison of Vehicle Detection Accuracy Between the Original Image Extent and the HAIR, and the Average Runtime of Using the HAIR on an Image from Each Camera

| Camera | Camera type | | Precision | | Recall | | RAP | | |
|---|---|---|---|---|---|---|---|---|---|
| | Height | Resolution | Image | HAIR | Image | HAIR | Image | HAIR | Runtime (s) |
| 1 | Low | Low | 0.987 | 0.964 | 0.724 | 0.938 | 0.722 | **0.885** | **0.0021** |
| 2 | High | Low | 0.993 | 1.000 | 0.209 | 0.717 | 0.271 | **0.727** | **0.0006** |
| 3 | Low | Low | 0.992 | 0.981 | 0.367 | 0.755 | 0.362 | **0.717** | **0.0005** |
| 4 | Low | Low | 0.992 | 0.982 | 0.772 | 0.969 | 0.722 | **0.895** | **0.0003** |
| 5 | Low | High | 0.938 | 0.929 | 0.830 | 0.970 | 0.781 | **0.852** | **0.0011** |
| 6 | High | High | 0.931 | 0.988 | 0.625 | 0.880 | 0.603 | **0.815** | **0.0003** |
| Overall | | | 0.972 | 0.974 | 0.588 | 0.871 | 0.577 | **0.815** | **0.0008** |

*Note:* HAIR = high-accuracy identification region; RAP = regional average precision.
The RAP and runtimes of using the HAIR are bolded.

**Table 3.** Comparison of Traffic Density Estimation Between the Original Image Extent and the HAIR

| Camera | Camera type | | Error | |
|---|---|---|---|---|
| | Height | Resolution | Image | HAIR |
| 1 | Low | Low | 3.858 | **1.760** |
| 2 | High | Low | 2.360 | **1.916** |
| 3 | Low | Low | 9.939 | **3.539** |
| 4 | Low | Low | 1.858 | **0.793** |
| 5 | Low | High | 2.210 | **2.102** |
| 6 | High | High | 0.733 | **0.603** |
| Overall | | | 3.493 | **1.785** |

*Note:* HAIR = high-accuracy identification region.
The errors of using the HAIR are bolded.

identify such areas is presented and tested. Our experiments show that the vehicle detection accuracy within the HAIR is significantly higher than that in the entire image extent (Table 2), suggesting that the HAIR consistently exists in images with the same field of view. Our results also indicate the effectiveness of the proposed algorithm, in which the desired accuracy of vehicle detection can be achieved within the HAIR. The accurate vehicle counts within the HAIR can subsequently be used to improve the accuracy of road traffic estimates obtainable from traffic camera images (Table 3).

The application of this algorithm partly relies on manual spatial data operations, such as georeferencing the images using ground control points. Although fixed traffic cameras are widely applied in traffic enforcement and monitoring (*42, 43*), applying this algorithm to real-time analysis on pan-tilt-zoom cameras will necessitate an automatic procedure so that the road traffic metrics derived from the HAIR can be estimated without labor-intensive human input. Some recent progress has the potential to address the issue of automation. High-quality spatial data have become increasingly available as they are used to enable the recent surge of geodesign (*44*). Road data in ArcGIS Urban, for example, have information about lanes and essential landmarks. Three-dimensional (3D) LiDAR data models are also used to automatically project two-dimensional (2D) traffic images into geographical space using deep neural networks (*45*). Finally, studies on automated landmark identification (*46*) make it possible to retrieve objects from traffic images that can be used for georeferencing. A fusion of these research outcomes is promising not only to improve the accuracy of automatic georeferencing, but also to automatically filter out the image regions where vehicles appear to be too small to be detected (*47*). This will significantly improve the automation of the applications on pan-tilt-zoom cameras.

The proposed algorithm has the potential to contribute to a broader field in big data-driven transport and urban analytics. The past two decades have witnessed the emergence of big mobility data contributed by the proliferating location-aware devices, such as various range sensors, GPS, and Wi-Fi (*48*), which have made it possible to address the data bottleneck for mobility research by providing large-scale human movement information conveniently and affordably (*49, 50*). However, a significant portion of the available big mobility data sets, such as mobile phone data and geo-tagged social media data, are retrieved from private data providers in which limited details of data production are released and only a few authorized users can gain access to the data (*51*). The lack of transparency in data production and the barriers in data sharing have added difficulties in the assessment of data reliability and quality and have subsequently limited the use of such data (*52, 53*). With the improvement in accuracy, the proposed algorithm can facilitate fully utilizing the new source of open geospatial big data, the publicly available traffic camera feeds, and contribute to the big data analytics for transportation and mobility research (*49, 54*). This algorithm can help extract accurate explicit movement information from open geospatial big data, and, more critically, provide reliable open mobility data products that have the potential to overcome the barriers in transparency and accessibility imposed by other data sources.

## Acknowledgments

Suggestions from all the anonymous reviewers are appreciated.

## Author Contributions

The authors confirm contribution to the paper as follows: study conception and design: Yue Lin, Ningchuan Xiao; data collection: Yue Lin; analysis and interpretation of results: Yue Lin; draft manuscript preparation: Yue Lin, Ningchuan Xiao. All authors reviewed the results and approved the final version of the manuscript.

## Declaration of Conflicting Interests

## Funding

The authors disclosed receipt of the following financial support for the research, authorship, and/or publication of this article: The authors acknowledge support from the Geospatial Fellows Program by the University of Illinois at Urbana-Champaign.

## ORCID iD

Yue Lin  https://orcid.org/0000-0001-8568-7734

## Supplemental Material

Supplemental material for this article is available online.